\documentclass[10pt,twocolumn,letterpaper]{article}

\usepackage{graphicx}
\usepackage{amsmath}
\usepackage{amssymb}
\usepackage{booktabs}
\usepackage{dsfont}

\newtheorem{definition}{Definition}[section]
\usepackage{multirow} 

\usepackage[pagebackref,breaklinks,colorlinks]{hyperref}

\usepackage[capitalize]{cleveref}
\crefname{section}{Sec.}{Secs.}
\Crefname{section}{Section}{Sections}
\Crefname{table}{Table}{Tables}
\crefname{table}{Tab.}{Tabs.}

\begin{document}

\title{
%Evaluating the Information Flows in the Explanation of Graph Neural Networks in Digital Pathology
Towards the Explanation of Graph Neural Networks in Digital Pathology with Information Flows
}

\author{Junchi Yu\\
Instittue of Automation, CAS\\
Beijing, China\\
\and
Tingyang Xu\\
Tencent AI Lab\\
Shenzhen, China\\
\and
Ran He\\
Instittue of Automation, CAS\\
Beijing, China\\
}
\maketitle

%%%%%%%%% ABSTRACT
\begin{abstract}
    %The wide applications of Graph Neural Networks (GNNs) in digital pathology lead to the 
    As Graph Neural Networks (GNNs) are widely adopted in digital pathology, there is increasing attention to developing explanation models (explainers) of GNNs for improved transparency in clinical decisions. 
    Existing explainers discover an explanatory subgraph relevant to the prediction. 
    However, such a subgraph is insufficient to reveal all the critical biological substructures for the prediction because the prediction will remain unchanged after removing that subgraph. 
    Hence, an explanatory subgraph should be not only necessary for prediction, but also sufficient to uncover the most predictive regions for the explanation.
    Such explanation requires a measurement of information transferred from different input subgraphs to the predictive output, which we define as information flow.
    In this work, we address these key challenges and propose IFEXPLAINER, which generates a necessary and sufficient explanation for GNNs.
    To evaluate the information flow within GNN's prediction, we first propose a novel notion of predictiveness, named $f$-information, which is directional and incorporates the realistic capacity of the GNN model.
    Based on it, IFEXPLAINER generates the explanatory subgraph with maximal information flow to the prediction.
    Meanwhile, it minimizes the information flow from the input to the predictive result after removing the explanation.
    Thus, the produced explanation is necessarily important to the prediction and sufficient to reveal the most crucial substructures.
    We evaluate IFEXPLAINER to interpret GNN's predictions on breast cancer subtyping.
    Experimental results on the BRACS dataset show the superior performance of the proposed method.

\end{abstract}

%%%%%%%%% BODY TEXT
\section{Introduction}

%-------------------------------------------------------------------------
%The emergence of deep learning has greatly facilitated histopathological image analysis. For example, Convolutional Neural Networks (CNNs) have been widely adopted in understanding digital pathology (DP) images to boost computer-aided diagnosis \cite{ballester2021artificial, browning2021digital}. 

%The emergence of Graph Neural Networks (GNNs) has greatly facilitated medical image analysis \cite{kipf2016semi}. 
%The message-passing scheme of GNNs is more efficient to accommodate the irregular and non-euclidean graph-structured data and exploits the relationship of biological entities, compared with the Convolutional Neural Networks (CNNs). 
%Hence, recent literature focus on employing GNNs to embed cells and cellular interactions in the RoIs of the digital pathological images to facilitate the computer-aided diagnosis, such as functional magnetic resonance image (fMRI) analysis \cite{li2021braingnn}, digital pathology \cite{jaume2021quantifying, jaume2020towards}, and precision oncology \cite{rhee2017hybrid,bera2019artificial}. 

Recent advances in deep learning have greatly boosted the development of histopathological image analysis \cite{ballester2021artificial, browning2021digital}. Especially, it becomes increasingly popular to leverage the Graph Neural Networks (GNNs) to exploit the complex relationship between the biological entities in digital pathology images \cite{kipf2016semi}. Various GNN-based methods are employed to facilitate clinical decisions such as cancer classification \cite{li2018graph,adnan2020representation, zhou2019cgc}, histopathological image segmentation \cite{zheng2019encoding,ozen2021self,graham2019hover} and cancer detection \cite{anklin2021learning}. 
%Especially, Graph Neural Networks (GNNs) have become a powerful tool  to, since the biological entities in the Region-of-Interests (RoIs) can be naturally represented as graph-structured data.
%Compared with Convolutional Neural Networks (CNNs), GNNs 
%The GNN-based methods have achieved superior performance in
Despite their success, the complex predictions made by GNNs are difficult for pathologists to understand \cite{jaume2021quantifying, sureka2020visualization}. 
The clinical decisions based on GNNs cannot be fully trusted by the practitioners without interpretations of the predictions of GNNs.
%the complex decision process of GNNs hinders the pathologist-intelligible explanations of their predictions. 
%The clinical decisions made by GNNs cannot be fully trusted and examined by the practitioners without understanding the predictions of GNNs. 
%Hence, it is essential to enable uncover the explainability of GNNs for pathologists.
%Therefore, it is imperative to investigate the explainability of GNNs within digital pathology \cite{ahmedt2021survey}.
Although many efforts have been made to increase the transparency of Convolutional Neural Networks (CNNs) \cite{binder2018towards,korbar2017looking,graziani2020concept,lu2021data}, explaining the predictions of GNNs in digital pathology is still a nascent research topic. 
%As GNNs employ the message-passing module to leverage the neighborhood information, it is essential to highlight the predictive compositions of the biological entities such as nuclei, cells and tissues. 
%Inspired by the explanation models (explaniers) for GNNs, several graph-pruning explainers are implemented to generate post-hoc explanations of the GNN-based cancer subtype classification \cite{jaume2020towards,jaume2021quantifying}.  

Inspired by the developments in the explainability of GNNs, several explanation models (explainers) are implemented to recognize the important subgraph of a biological entity-graph that is predictive to the GNN's prediction \cite{jaume2020towards}.
However, the discovered subgraph is {\em insufficient} to fully uncover the explainability of GNNs since the GNNs probably make the same prediction after removing the subgraph. 
This indicates that these explainers potentially ignore some informative substructures to the prediction. 
Hence, apart from discovering an important subgraph for the explanation, explainers are also supposed to investigate whether removing the explanatory subgraph sufficiently leads to a different prediction.
%, which will be attributed to answer such counterfactual problem: 
%"\textit{Will the GNN still make the same prediction if the explanatory subgraph is removed?}"
The recently proposed counterfactual-based explainers \cite{lucic2021cf} partially address this issue by recognizing the minimal explanatory subgraph, if removed, that leads to an alternative prediction.
However, the found subgraph may contain {\em unnecessarily} significant portion to the prediction.
Suffering from the above two issues for recent GNN explainers, it motivates us to generate an ideal explainable subgraph which is not only {\em necessary} for GNNs to make a specific prediction, but also {\em sufficient} to uncover most crucial regions for the explanation.

%Generating such an explanation requires quantifying the information transferred from different input substructures to the GNN's prediction, which is nontrivial. 
%Generating such an explanation requires quantifying the information transferred from different input substructures to the GNN's prediction, which is defined as the information flow. 

Generating such an explanation requires identifying the contributions of different input substructures to the GNN's prediction.
To articulate this concept, we define the information transferred from the input to the prediction through the GNN models as the information flow. 
The information flow within GNNs is directional since we can only infer the underlying label of the input instead of vice versa.
Moreover, it relies on a realistic capacity of GNN models to describe the information transfer.
For example, a more powerful GNN predicts more easily than a less powerful one, which indicates the information flow varies in different GNN models.
Existing methods \cite{yu2020graph,ying2019gnnexplainer} tend to measure such information flow with mutual information.
However, the mutual information is symmetric and only measures relevance between the input and its prediction without considering the modeling power of the GNN.

%Existing notions of informativeness, such as mutual information, is less appropriate to evaluate such information flows.
%First, the mutual information is symmetric, which means the input and its prediction are equally informative to each other \cite{xu2020theory}.
%However, the information flow within GNNs is directional since we can only infer the underlying label of the input instead of vice versa.
%Moreover, the mutual information does not consider the realistic capability of the GNNs.
%For example, a more powerful and expressive GNN can predict more easily than a less powerful one. 
%Thus, the information flow from the same input and its prediction varies in different GNN models.

%Based on the recent advances in machine learning \cite{xu2020theory}, we introduce a new notion of information, namely $f$-Information, to evaluate the information flow under the realistic capacity of GNN models.
To measure the information flow within GNN's prediction, we introduce a new notion of informativeness, namely $f$-Information, to evaluate the information flow within GNN's prediction.
Unlike mutual information, $f$-Information is directional and incorporates the realistic capacity of the GNN models.
Hence, we can identify the significance of different input substructures to the prediction by evaluating the corresponding information flows with $f$-Information.
Based on $f$-Information, we emphasize the sufficient and necessary nature of the explanation and propose a novel explanation model, namely IFEXPLAINER, to faithfully explain the GNNs in digital pathology tasks.
Specifically, it maximizes the information flow from the explanatory subgraph to the prediction while minimizing the information flow from the input to the prediction after removing the explanatory subgraph.
The produced explanatory subgraph is thus necessarily important for the GNN to make the prediction and sufficiently change the prediction if removed.
Therefore, the generated explanation faithfully interprets the GNN's prediction by uncovering the most crucial substructures.
Moreover, we introduce a tractable optimization scheme for IFEXPLAINER to explain the predictions of biological instances efficiently.
Furthermore, to comprehensively evaluate the explainers in terms of necessity and sufficiency, we propose a set of pathologist-intelligible metrics based on subgraph-level separability.
We employ IFEXPLAINER to explain the predictions of the GNN on breast cancer subtyping.
Qualitative and quantitative results on BRACS dataset \cite{pati2020hact} show the superior performance of the proposed method.

The contributions of this work are summarized as follows:
\begin{itemize}
    \item We propose a novel notion of $f$-Information to measure the information flows in the GNN's prediction.
    \item We emphasize the necessary and sufficient nature of the explanation and propose the IFEXPLAINER to faithfully interpret GNN's predictions by evaluating the information flows.
    \item We introduce a set of evaluation metrics based on subgraph level separability for the pathologist-intelligible evaluations.
    \item Extensive studies on the BRACS dataset verify the superior performance of the proposed method.
\end{itemize}

\section{Related work}
\subsection{GNNs in Digital Pathology}
%The graph-structured data such as the cell graphs are wildly adopted to model the morphological and topological information of cells and cellular interactions in digital pathology.
%Compared with the histopathology images, the formulation of cell graphs directly embed the cells in RoIs at the entity level and is more easily understood by pathologists.
%Thus, GNN-based models become popular to analyze the cell graphs in digital pathology. 
%Recent studies represent the RoIs of histopathological images as cell graphs to better exploit the multi-attribute and topological information of cells and cellular interaction \cite{ahmedt2021survey}. 
Recent studies employ graph techniques to embed the biological entities such as cells, tissue regions, and patches into graph-structured data \cite{ahmedt2021survey, jaume2021histocartography}.
Hence, GNN-based methods become popular to analyze such graph-structured data to facilitate medical decisions.
A large body of these works focuses on the application of GNNs on cancer subtype classification, such as breast cancer classification \cite{rhee2017hybrid, anand2020histographs, jaume2020towards}, colorectal cancer classification \cite{zhou2019cgc,studer2021classification} and lung cancer classification \cite{li2018graph,adnan2020representation}. For example, the attention-based robust spatial filtering method is adopted to highlight the important cells to enhance the GNN's performance on breast cancer subtype classification \cite{sureka2020visualization}. To exploit the hierarchical graph structure, Hact-net builds the tissue graph and cell graph to capture the cellular attributes at different levels \cite{pati2020hact}. Slide-Graph enriches the topological context of nodes via aggregating the local features in the cell graph \cite{lu2020capturing}. Apart from the applications in cancer subtype classification, other works leverage GNN for histopathological image segmentation \cite{anklin2021learning} and cancer detection \cite{ozen2021self}.  GNN is also cooperated with the contrastive predictive coding for cancer detection in a weakly supervised manner using only labels at the tissue micro-arrays level \cite{wang2020weakly}. While GNN has made tremendous progress in digital pathology, the explainability of GNN's prediction is less explored. Several explanation methods of GNN are implemented to digital-pathology tasks \cite{jaume2021quantifying}. However, the generated explanations do not uncover all the important substructures since they are insufficient to change the predictions if removed. 

\subsection{Explanability of GNNs}
While the advancements in GNNs have revolutionized deep graph learning, the explainability of GNNs is lagged \cite{ying2019gnnexplainer}. 
The discrete and combinatorial nature of graph-structured data hinders the development of GNNs' explanation models \cite{yuan2021explainability,yu2020graph}. To this end, several methods have been proposed to discover an important subgraph of the input for the explanation \cite{ying2019gnnexplainer,luo2020parameterized,yu2021recognizing,huang2020graphlime}. The gradient-based methods such as GraphGRADCAM \cite{pope2019explainability} and GraphGRAD-CAM++ \cite{jaume2021quantifying} employ gradient values to evaluate the importance of each node based on CAM \cite{selvaraju2017grad}. Similarly, GraphLRP decomposes the prediction into several terms concerning the nodes and edges \cite{schwarzenberg2019layerwise}. Recently, it becomes popular to leverage the graph-pruning explainers to discover an explanatory subgraph that is most relevant to the static prediction. GNNEXPLAINER generates post-hoc explanations by maximizing the mutual information between the subgraphs and GNN's predictions \cite{ying2019gnnexplainer}. Furthermore, PGEXPLAINER learns a parametric model to generate explanations \cite{luo2020parameterized}. Similar to PGEXPLAINER, GraphMask generates the explanatory subgraph in a layer-wise manner \cite{schlichtkrull2020interpreting}. Although these methods can generate explanations in the inductive setting, they heavily rely on the node embeddings from the GNNs \cite{lin2021generative}, thus lacking the understanding of the biological graphs in digital pathology. The explainers such as SubgraphX \cite{yuan2021explainability} and XGNN \cite{yuan2020xgnn} formulate the generation of explanations as a reinforcement learning task. However, it is time-consuming to verify the contribution of each node for large-scale biological graphs via reinforcement learning.
While the graph-pruning explainers explain the individual predictions, the counterfactual-based explainers \cite{lucic2021cf,bajaj2021robust} recognize a minimal subgraph, if moved, can lead to the drastic change in GNN's prediction. While the discovered subgraph sufficiently changes the prediction if removed, it is may contain unnecessary important portions to the prediction.

\section{Methodology}
%In this section, we begin with an brief introduction to the cancer subtype classification with GNN. Then, we propose a causal framework to unify various explanation models of GNN and propose a novel approach, namely PAEXPLAINER, to efficiently discover causal explanations of GNN's decisions.
In this section, we begin with the formulation of the information flow as $f$-information. Then we introduce how information flows are used to comprehensively explain GNNs as IFEXPLAINER and how to optimize IFEXPLAINER. Finally, we discuss a novel evaluation metrics to measure how necessary and sufficient the selected subgraphs make the explanations of GNNs.

\subsection{$f$-Information}
Given two random variables $X$ and $Y$, the mutual information $I(X,Y)$ is defined as follows:
%The mutual information $I(X,Y)$ measures the relevance between two random variables $X$ and $Y$:
\begin{equation}
\begin{aligned}
I(X,Y)=\iint_{x,y}p(x,y)\log{\frac{p(x,y)}{p(x)p(y)}}\mathrm{d}x\mathrm{d}y
\end{aligned}
\end{equation}
The mutual information is widely adopted to measure the relevance between random variables in representation learning. 
However, it is less appropriate to reason the explainability of GNNs with the mutual information. 
To explain GNN's predictions, it is vital to investigate the information flow with GNNs, which is defined as the information transferred from the input to the output of GNNs.
However, the mutual information is insufficient to represent such directional information flows. 
Specifically, the mutual information $I(X,Y)$ is symmetric, which means $X$ and $Y$ exchange equal information. 
This does not hold in GNN's prediction since we can only infer the prediction $Y$ from the input $X$ instead of vice versa \cite{xu2020theory}, which indicates that the information flow is directional.
Moreover, the formulation of mutual information does not consider the modeling power of GNNs.
In general, a more powerful GNN can infer $Y$ from $X$ more easily than a less powerful one, which shows the information flow from $X$ to $Y$ varies in different GNN models.
The above analysis motivates us to find a proper metric to quantify the directional information flows in GNN's prediction.
%to quantify the information flows between the input graphs and the output prediction in the explainability of GNNs. 
%First, the formulation of mutual information does not consider the modeling power of the classifier \cite{xu2020theory}. 
%In general, a more powerful classifier can infer $Y$ from $X$ more easily than a less powerful one, which shows the information flow from $X$ to $Y$ varies in different classifier.
%Second, the symmetric nature of mutual information is less appropriate to model asymmetric information flows within the classifier. 
%For example, given a classifier, it is easier to infer the prediction $Y$ of the input $X$ than vice versa. 
%This shows the information flow from $X$ to $Y$ is larger than that from $Y$ to $X$.
%As the classifier is alway one-way, it is essential to 
%quantify how the input is predictive to the prediction given the modeling power of a classifier.
\begin{figure}[t]
\begin{center}
%\framebox[4.0in]{$\;$}
\centerline{\includegraphics[width=1.0\columnwidth]{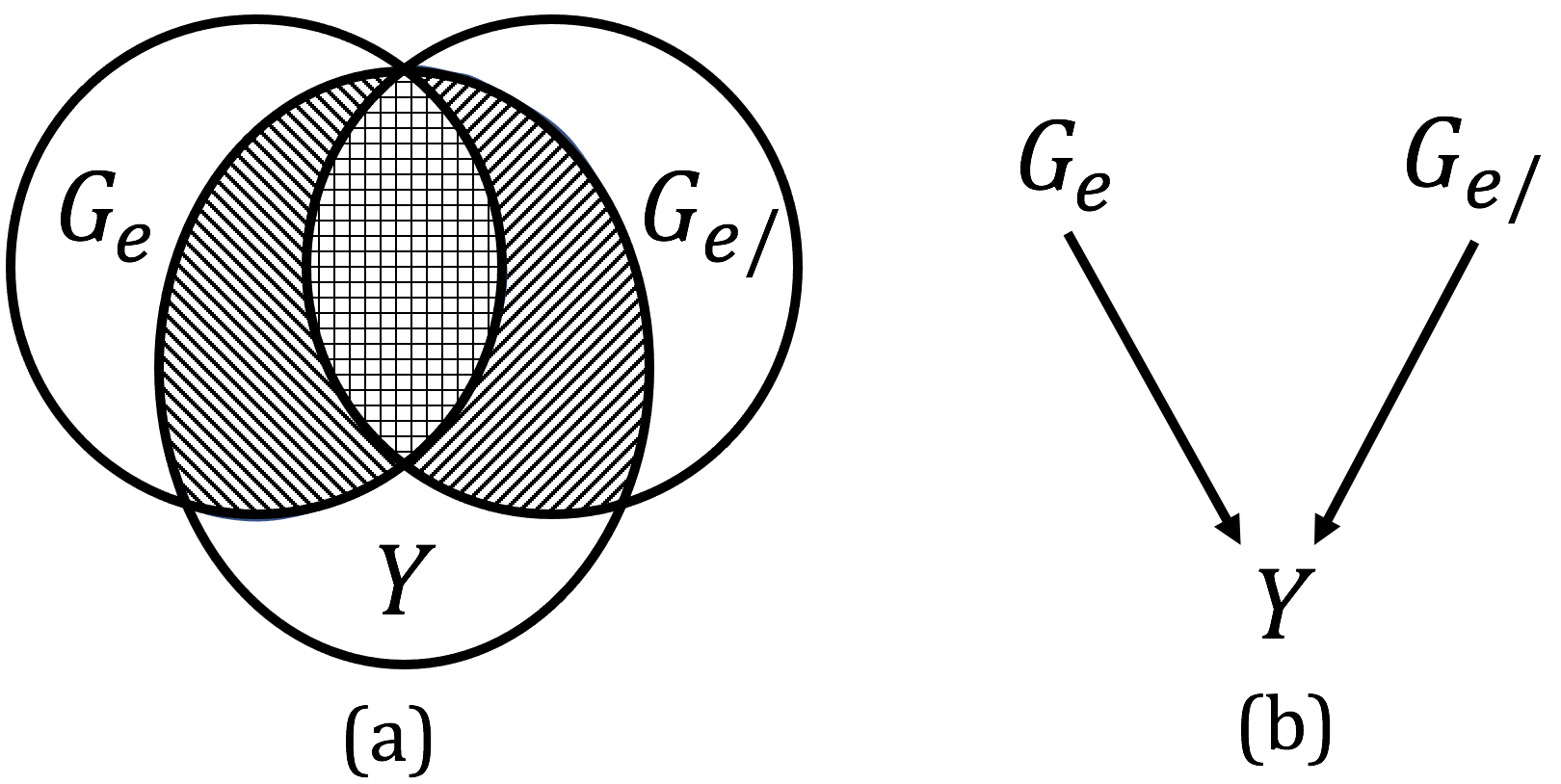}}
\vspace{-1cm}
\end{center}
\caption{(a). The Venn diagram illustrating the relationship between the explanatory subgraph $G_{e}$, the complementary subgraph $G_{e\backslash}$ and the prediction $Y$. The circles of $G_{e}$ and $G_{e\backslash}$ denote the entropy while the circle of $Y$ is the $f$-entropy. The intersections of two circles represent $I_{f}(G_{e}\rightarrow Y)$ and $I_{f}(G_{e\backslash}\rightarrow Y)$. When $Y$ is known, the intersection of $G_{e}$ and $G_{e\backslash}$ vanishes due to the assumption that $G_{e}$ and $G_{e\backslash}$ are conditionally independent given $Y$.  (b). The graphic model of IFEXPLAINER. The arrows represent directions of $f$-Information.} 
\label{veen}
\vspace{-0.5cm}
\end{figure}
Inspired by recent advances in machine learning \cite{xu2020theory}, we propose a novel $f$-Information to exploit the information flow. 
\begin{definition}
Given $X$ and $Y$ be the input graph and the corresponding prediction, and $f$ is the GNN that maps $X$ to $Y$. The $f$-Information $I_{f}(X\rightarrow Y)$ is defined as follows:
\begin{equation}
\begin{aligned}
H_{f}(Y)&=\mathrm{E}_{y}-\log f(y)\\
H_{f}(Y|X)&=\mathrm{E}_{x,y}-\log f[x](y)\\
I_{f}(X\rightarrow Y) &= H_{f}(Y)-H_{f}(Y|X),
\end{aligned}
\label{f-info}
\end{equation}
where $f[x](y)$ is the output distribution of $y$ after $f$ receives the input $x$ and $f(y)=\mathrm{E}_{x}f[x](y)$ is the marginalized distribution of $y$.
$H_{f}(Y)$ and $H_{f}(Y|X)$ are the $f$-entropy and conditional $f$-entropy. $I_{f}(X\rightarrow Y)$ is the $f$-Information from $X$ to $Y$ and the arrow denotes the direction.
\end{definition}
Similar to the predictive $\mathcal{V}$-Information \cite{xu2020theory}, $f$-Information is a variational extension to the Shannon mutual information by incorporating the modeling power of the classifier. 
Differently, $f$-Information measures the information flow in a specific GNN while the $\mathcal{V}$-information focuses on learning optimal representations within a predictive family. 
Notice that $f$ in Eq.~\ref{f-info} can be replaced by other neural networks such as CNN and RNN, and thus permits the application of $f$-Information to the explainability of other models.
In our work, we focus on the explainability of GNNs in digital pathology.

%Similar to the predictive $\mathcal{V}$-Information \cite{xu2020theory}, the $f$-Information is directional and asymmetric. Hence, we can investigate the information flow from the input to the output of the predictor. Differently, the $f$-Information is defined over a fixed predictor rather than a predictive family. Thus, it is more convenient to reason the explainability of a pretrained classifier with the $f$-Information. 
%Notice that $f$ can be any neural networks such as CNN, RNN and GNN. In our work, we consider the explainability of GNNs in digital pathology. 

\subsection{Information Flow in GNN's Explanation}
%The explanation models of GNNs generally recognize the subgraph of the input which is most informative to the prediction of GNNs. We formulate this problem as recognizing the subgraph of input, which has maximal information flow to the prediction with the notion of $f$-Information.
\begin{figure*}[t]
\begin{center}
%\framebox[4.0in]{$\;$}
\centerline{\includegraphics[width=2.0\columnwidth]{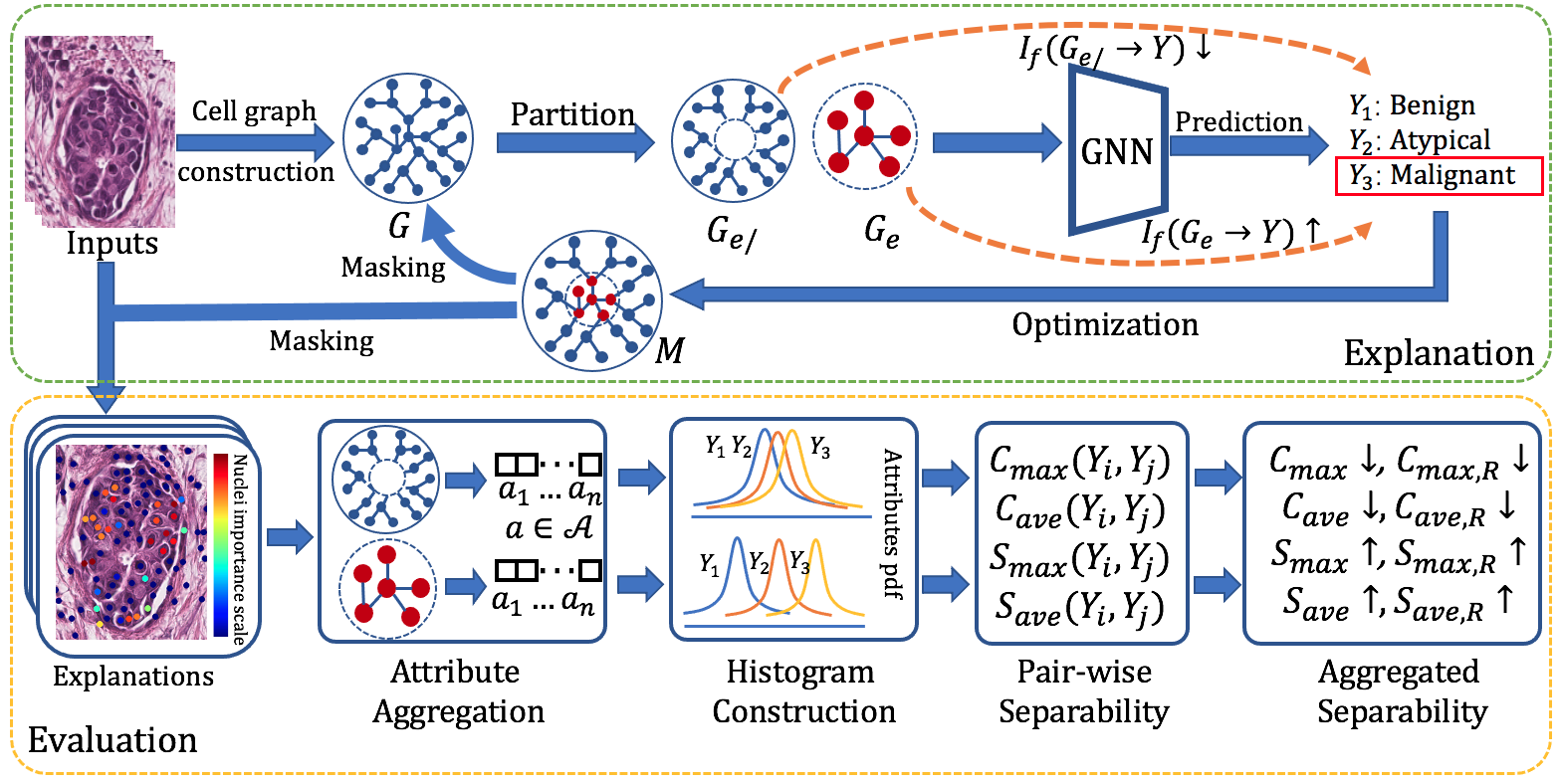}}
\end{center}
\vspace{-0.8cm}
\caption{The framework of the proposed method. IFEXPLAINER generates necessary and sufficient explanations of GNNs by contradicting the information flows $I_{f}(G_{e}\rightarrow Y)$ and $I_{f}(G_{e\backslash}\rightarrow Y)$. To evaluate the necessity and sufficiency of the explanations, we aggregate the pathologist-intelligible attributes of $G_{e}$ and $G_{e\backslash}$ respectively and calculate their inter-class separability.} 
\label{flowchart}
\vspace{-0.5cm}
\end{figure*}
A faithful explanation is supposed to be \textit{necessary}  for GNNs to make the prediction and  \textit{sufficient} to uncover most predictive regions.
We formulate these conditions by measuring the corresponding information flows.
Specifically, we recognize the explanatory subgraph of the input, which transfers maximal information flow to the GNN's prediction. Meanwhile, we restrict the information flow from the input to the prediction after removing the explanatory subgraph.
%Different from the existing graph-pruning explainers \cite{jaume2021histocartography,jaume2020towards}, we also investigate the changes of predictions after removing the explanatory subgraphs to interpret the dynamics of GNN's prediction.

 %We make the assumption that $G_{e}$ and $G_{e\backslash}$ are conditionally independent given $Y$, under the intuition that 
%Notice that different graphs may contain different subgraphs. Hence, the distribution of $G_{e}$ and $G_{e\backslash}$ are as follows:
%\begin{equation}
%\begin{aligned}
%p(G_{e})=\mathrm{E}_{G\in p(G)} P(G_{e}|G),
%p(G_{e\backslash})=\mathrm{E}_{G\in p(G)} P(G_{e\backslash}|G).
%\end{aligned}
%\end{equation}

Let $G\in \mathcal{G}$ and $Y\in \mathcal{Y}$ be the biological entity-graph such as the cell graph and the GNN's prediction. 
$f$ is the GNN to be explained, which maps $G$ to $Y$.
Define $G_{e}$ and $G_{e\backslash}$ as the explanatory subgraph and the complementary subgraph after removing $G_{e}$ from $G$. 
The relationship of $G_{e}$, $G_{e\backslash}$ and $Y$ are shown in Figure~\ref{veen} (a). 
In Figure~\ref{veen} (a), the circle of $Y$ represents the $H_{f}(Y)$. 
The circles of $G_{e}$ and $G_{e\backslash}$ are $H(G_{e})$ and $H(G_{e\backslash})$.
And intersections of two circles denote the information flows which are measured by $I_{f}(G_{e}\rightarrow Y)$ and $I_{f}(G_{e\backslash}\rightarrow Y)$ respectively. 
As shown in Figure~\ref{veen}, $Y$ receives information from both $G_{e}$ and $G_{e\backslash}$ since the GNN $f$ takes the whole structure of $G$ to make the prediction $Y$.
To produce a faithful explanation of the GNN, we need to maximize the information flow from $G_{e}$ to $Y$, which is equal to maximize $I_{f}(G_{e}\rightarrow Y)$.
Moreover, after removing $G_{e}$ from $G$, $Y$ only receives the information flow from $G_{e\backslash}$.
$G_{e\backslash}$ is less predictive to $Y$,
as we hope $G_{e}$ that reveals all the information for predicting $Y$.
Intuitively, we hope the GNN will make a different prediction after removing $G_{e}$ from $G$, which requires to minimize $I_{f}(G_{e\backslash}\rightarrow Y)$. 
%In Figure~\ref{veen} (a), the intersection of two circles represent the $f$-Information of two random variables.
%Hence, the intersection between the circles of $Y$ and $G_{e}$ is $I_{f}(G_{e}\rightarrow Y)$, and the overlap between the circles of $Y$ and $G_{e\backslash}$ is $I_{f}(G_{e\backslash}\rightarrow Y)$. 
The above analysis leads to a novel explainer, namely IFEXPLAINER, which leverages the \textbf{I}nformation \textbf{F}lows from different input substructures to the GNN's predictions. 
We present the graphical model of IFEXPLAINER in Figure~\ref{veen} (b) and the arrows indicate the directions of the information flows. 
The framework of IFEXPLAINER is shown in Figure~\ref{flowchart}.
%The graphical model of IFEXPLAINER is shown in figure~\ref{veen} (b) and the arrows indicate the directions of the information flows. 
The objective of IFEXPLAINER is as follows:
\begin{equation}
\begin{aligned}
&\max_{p_{\theta}(G_{e})}I_{f}(G_{e}\rightarrow Y)-I_{f}(G_{e\backslash}\rightarrow Y)\\
&s.t. |G_{e}|\leq K,
\end{aligned}
\label{og-obj}
\end{equation}
where $p_{\theta}(G_{e})=\int_{G} p_{\theta}(G_{e}|G)p(G)\mathrm{d}G$ is the parametric distribution of $G_{e}$ and $|\cdot|$ denotes the number of nodes in $G_{e}$. We introduce a constant $K$ to constrain the size of $G_{e}$, and thus prevent from the trivial solution that $G_{e}=G$ and $G_{e\backslash}=\phi$. By introducing a Lagrange Multiplier $\beta$, we convert the constrained optimization problem into an unconstrained one:
\begin{equation}
\begin{aligned}
&\min_{G_{e}\sim p_{\theta}(G_{e})}-I_{f}(G_{e}\rightarrow Y)+I_{f}(G_{e\backslash}\rightarrow Y)+\beta |G_{e}|.
\end{aligned}
\label{lm-obj}
\end{equation}
%The objective of IFEXPLAINER in Eq.~\ref{lm-obj} is related to the Information Bottleneck with Side Information (IBSI) by treating $G_{e\backslash}$ as the side information variable. 
%The IBSI seeks for a compressed representation $Z$ of the input $X$, which is maximally informative to the positive label $Y^{+}$ and minimally informative to the negative label $Y^{-}$:
%\begin{equation}
%\begin{aligned}
%&\min_{Z} -I(Y^{+},Z) + I(Y^{-},Z)+\beta I(X,Z).
%\end{aligned}
%\label{IBSI-obj}
%\end{equation}
%Analy $G_{e\backslash}$ 

%Both objectives in Eq.~\ref{lm-obj} and Eq.~\ref{IBSI-obj} seek for a compressed structure of the input. Differently, IFEXPLAINER employs the asymmetric $f$-Information to measure the information flow from different partitions of the input graphs to the outputs of GNN, while IBSI use the symmetric mutual information to quantify the relevance of the latent variable to different labels. 
%Moreover, IBSI  IFEXPLAINER compresses for 

\subsection{Optimization Scheme for IFEXPLAINER}
%Optimizing the objective in Eq.~\ref{lm-obj} is intractable, mostly due to the discrete and irregular nature of the graph-structured data. 
We introduce the optimization Scheme for minimizing the objective of IFEXPLAINER in Eq.~\ref{lm-obj}.  We first examine the $f$-Information terms in the objective:
\begin{equation}
\begin{aligned}
\mathcal{L}_{f}&=-I_{f}(G_{e}\rightarrow Y)+I_{f}(G_{e\backslash}\rightarrow Y)\\
&= H_{f}(Y|G_{e})-H_{f}(Y|G_{e\backslash})
\end{aligned}
\label{info-loss}
\end{equation}

\begin{figure*}[t]
\begin{center}
%\framebox[4.0in]{$\;$}
\centerline{\includegraphics[width=2.0\columnwidth]{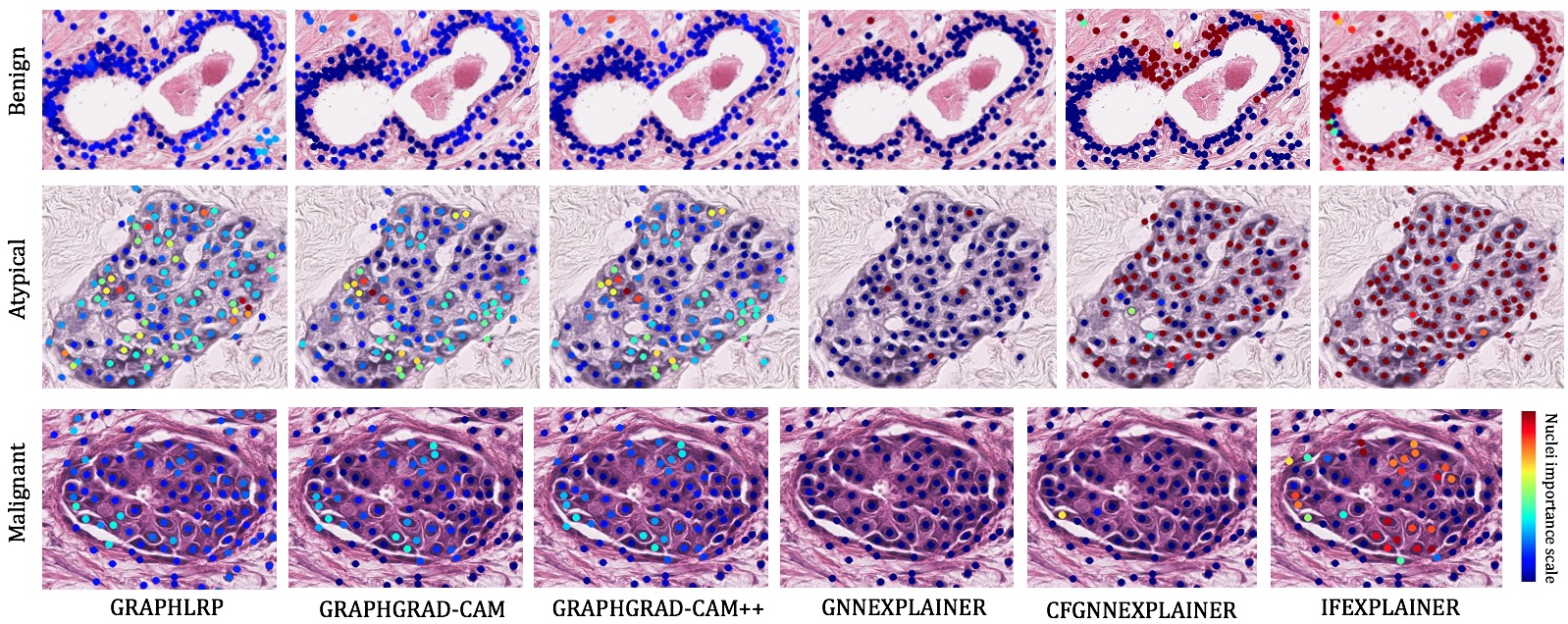}}
\vspace{-1.1cm}
\end{center}
\caption{Explanations for the breast cancer subtyping. The rows and columns represent cancer subtypes and the explanation results of different explainers. The important nuclei are highlighted according to their importance scores.} 
\label{show-cells}
\vspace{-0.7cm}
\end{figure*}
%We simply the to generate a post-hoc explanation to the prediction of the GNN model.
\textbf{Single Instance Explanation:} The conditional $f$-entropy can be estimated by the batched data $\{(G_{i},Y_{i})|i=1,\cdots,N\}$ with provable guarantees. Please refer to Supplementary Materials for more details. Let $G_{e,i}\sim p_{\theta}(G_{e}|G_{i})$ be the explanatory subgraph of $G_{i}$ and $G_{e\backslash,i}$ is the corresponding complementary subgraph. We simplify the loss term in Eq.~\ref{info-loss} as follows:
\begin{equation}
\begin{aligned}
\mathcal{L}_{f}&=\frac{1}{N}\sum_{i=1}^{N}\log{f[G_{e\backslash,i}](Y_{i})}-\log{f[G_{e,i}](Y_{i})}.
\end{aligned}
\label{info-loss-sim}
\end{equation}
Notice that $G_{i}\perp G_{j}$ for $i\neq j$, we plug Eq.~\ref{info-loss-sim} into Eq.~\ref{lm-obj} and further simplify the objective for the single-instance explanation:
\begin{equation}
\begin{aligned}
\mathcal{L}&=\log{f[G_{e\backslash,i}](Y_{i})}-\log{f[G_{e,i}](Y_{i})} + \beta |G_{e,i}|\\
&=\mathcal{L}_{f}+\beta |G_{e,i}|.
\end{aligned}
\label{single}
\end{equation}
As $f[G_{e\backslash,i}](Y_{i})\in [0,1]$, minimizing $\log{f[G_{e\backslash,i}](Y_{i})}$ probably results in the negative infinity , leading to an unstable training process. Thus, we replace this term with a minimax loss:
\begin{equation}
\begin{aligned}
\mathcal{L}_{f} =& \max[\log{f[G_{e\backslash,i}](Y_{i})},\min_{j\neq i}(\log{f[G_{e\backslash,i}](Y_{j})})]\\
&-\log{f[G_{e,i}](Y_{i})},
\end{aligned}
\label{minimax}
\end{equation}

\textbf{Continuous Relaxation.} However, it is still intractable to minimize the objective in Eq.~\ref{single}, mostly due to the discrete and irregular nature of the graph-structured data. Moreover, sampling the explanatory subgraph is computationally inefficient since a graph has exponentially many subgraphs.
Thus, we consider learning a differentiable soft mask to generate the explanatory subgraph.
Different from the practice in GNNEXPLAINER \cite{ying2019gnnexplainer}, we do not mask node features since we only highlight the critical topology of the biological graph to GNN's prediction, which is more intuitive and human-intelligible \cite{yuan2021explainability}. 
Given an input graph $G$ with $n$ nodes, we obtain $G_{e}$ by applying a node mask $M\in \mathcal{R}^{n}$ to $G$. The elements of $M$ are bounded in $[0,1]$, which indicates the corresponding nodes' importance scores.
And we mask $G$ with $1-M$ to obtain $G_{e\backslash}$. 
%\begin{equation}
%\begin{aligned}
%&M_{V} = [\underbrace{\sigma(M),\cdots,\sigma(M)}_{d}], M_{A} = [\underbrace{\sigma(M),\cdots,\sigma(M)}_{n}],\\
%&V_{e} = V \odot M_{V}, A_{e} = A \odot M_{A}\odot M_{A}^{T},
%\end{aligned}
%\label{mask}
%\end{equation}
%where $[\cdots]$ is the concatenation operation to generate node mask $M_{V}\in\mathcal{R}^{n\times d}$ and adjacent mask $M_{A}\in\mathcal{R}^{n\times n}$. $\sigma$ denotes the sigmoid function which map the values in $M$ into $[0,1]$ and $\odot$ is the element-wise Hadamard product.
Note that $M$ is a soft mask with continues values instead of a binary mask, we further employ an entropy constraint to encourage discrete values in $M$ \cite{luo2020parameterized}:
\begin{equation}
\begin{aligned}
\mathcal{L}_{ent} = \sum_{i=1}^{n} -M_{i}\log{M_{i}} -(1-M_{i})\log (1-M_{i}),
\end{aligned}
\label{entropy}
\end{equation}
where $M_{i}$ is the i-th element of $M$. We further convert the size constraint $|G_{e}|$ in Eq.~\ref{single} to the constraint on the summation of the elements in $M$: $\mathcal{L}_{size} =|G_{e}|= \sum_{i=1}^{n} M_{i}$.
%\begin{equation}
%\begin{aligned}
%\mathcal{L}_{size} = \sum_{i=1}^{n} M_{i},
%\end{aligned}
%\label{size}
%\end{equation}
Thus, the total loss function of IFEXPLAINER is:
\begin{equation}
\begin{aligned}
\mathcal{L} = \mathcal{L}_{f} +\beta\mathcal{L}_{size} + \gamma \mathcal{L}_{ent},
\end{aligned}
\label{total-objective}
\end{equation}
where $\beta$ and $\gamma$ are hyper-parameters.

\begin{table*}[t]
  \centering
  \caption{Quantitative results of different explainers. We compare the average and maximal metrics of the pair-wise and aggregated separability scores.}
  \vspace{-0.35cm}
  \footnotesize
  \setlength{\tabcolsep}{1.15mm}{
    \begin{tabular}{c|c|c|c|c|c|c|c|c|c|c}
    \toprule
    Class pair & \multicolumn{2}{c|}{B vs. A} & \multicolumn{2}{c|}{B vs. M} & \multicolumn{2}{c|}{A vs. M} & \multicolumn{4}{c}{B vs. A vs. M} \\
    \midrule
    Metric & $C_{max}\downarrow$ & $S_{max}\uparrow$ & $C_{max}\downarrow$ & $S_{max}\uparrow$ & $C_{max}\downarrow$ & $S_{max}\uparrow$ & $C_{max}\downarrow$ & $S_{max}\uparrow$ & $C_{max,R}\downarrow$ & $S_{max,R}\uparrow$ \\
    \midrule
    GRAPHGRAD-CAM & 0.393 & 0.328 & 0.310  & 0.453 & 0.477 & \underline{0.598} & 1.180  & 1.379 & 1.489 & 1.832 \\
    GRAPHGRAD-CAM++ & 0.404 & 0.318 & \underline{0.307} & 0.457 & 0.477 & \textbf{0.619} & 1.188 & \underline{1.394} & 1.495 & \underline{1.850} \\
    GRAPHLRP & 0.360 & 0.268 & 0.310  & 0.313 & 0.610  & 0.310  & 1.28  & 0.892 & 1.590  & 1.205 \\
    GNNEXPLAINER & 0.373 & 0.307 & 0.285 & 0.479 & 0.536 & 0.401 & 1.194 & 1.187 & 1.480  & 1.666 \\
    CFGNNEXPLAINER & 0.350 & 0.226 & 0.304 & 0.514 & 0.475 & 0.41 & 1.129 & 1.15 & 1.433  & 1.664 \\
    \textbf{IFExplainer}    & \underline{0.338} & \textbf{0.355} & \textbf{0.290} & \textbf{0.521} & \textbf{0.465} & 0.542 & \textbf{1.093} & \textbf{1.418} & \textbf{1.383} & \textbf{1.939} \\
    \textbf{IFExplainer w/o $\mathcal{L}_{size}$} & 0.378 & \underline{0.345} & 0.319 & 0.492 & 0.519 & 0.407 & 1.216 & 1.244 & 1.534 & 1.736 \\
    \textbf{IFExplainer w/o $\mathcal{L}_{ent}$} & \textbf{0.336} & 0.317 & 0.322 & \underline{0.510}  & \underline{0.474} & 0.467 & \underline{1.121} & 1.294 & \underline{1.443} & 1.803 \\
    \midrule
    Metric & $C_{ave}\downarrow$ & $S_{ave}\uparrow$ & $C_{ave}\downarrow$ & $S_{ave}\uparrow$ & $C_{ave}\downarrow$ & $S_{ave}\uparrow$ & $C_{ave}\downarrow$ & $S_{ave}\uparrow$ & $C_{ave,R}\downarrow$ & $S_{ave,R}\uparrow$ \\
    \midrule
    GRAPHGRAD-CAM & 0.272 & 0.247 & 0.212 & 0.302 & 0.319 & \underline{0.335} & 0.803 & 0.884 & 1.015 & 1.187 \\
    GRAPHGRAD-CAM++ & 0.276 & 0.249 & \underline{0.208} & 0.306 & 0.319 & \textbf{0.343} & 0.803 & 0.897 & 1.011 & 1.203 \\
    GRAPHLRP & 0.253 & 0.227 & 0.257 & 0.197 & 0.356 & 0.235 & 0.866 & 0.659 & 1.123 & 0.857 \\
    GNNEXPLAINER & \textbf{0.241} & 0.240  & 0.249 & \underline{0.343} & 0.323 & 0.32  & 0.813 & 0.903 & 1.062 & 1.247 \\
    CFGNNEXPLAINER & 0.258 & 0.188 & 0.213 & 0.338 & 0.297 & 0.277 & 0.768 & 0.803 & 0.981  & 1.141 \\
    \textbf{IFExplainer}  & 0.251 & \textbf{0.257} & \textbf{0.207} & \textbf{0.338} & \textbf{0.290} & 0.327 & \textbf{0.748} & \textbf{0.922} & \textbf{0.955} & \textbf{1.261} \\
    \textbf{IFExplainer w/o $\mathcal{L}_{size}$} & \underline{0.246} & \underline{0.253} & 0.270  & 0.273 & 0.343 & 0.327 & 0.858 & 0.854 & 1.127 & 1.127 \\
    \textbf{IFExplainer w/o $\mathcal{L}_{ent}$}  & 0.263 & 0.246 & 0.216 & 0.341 & \underline{0.296} & 0.325 & \underline{0.775} & \underline{0.912} & \underline{0.991} & \underline{1.253} \\
    \bottomrule
    \end{tabular}}%
  \label{separability}%
  \vspace{-0.7cm}
\end{table*}%

\subsection{Subgraph-level Evaluation Metrics}
\label{section-sep}
We propose a set of pathologist-intelligible evaluation metrics for the explanations of GNNs on breast cancer subtyping with the relevant pathological attributes. 
Our evaluation metrics aggregate the pathological attributes of the nuclei for the explanatory and complementary subgraphs, and evaluate their inter-class separability scores at the subgraph level.
Hence, they provide comprehensive assessments considering the necessary and sufficient nature of the explanations. 
And they are less affected by the diverse numbers of nuclei in different cell graphs.

%considering the  sufficiency and necessity of the explanation.
%Different from prior work \cite{jaume2021histocartography} which computes separability scores of the important nuclei, our evaluation metrics are based on the inter-class separability scores at subgraph level.
%Hence, our evaluation metrics are less affected by the diverse numbers of nuclei in different cell graphs.
%We further evaluate the dynamic interpretations of GNNs with the separability scores of the complementary subgraphs, which are obtained by removing the explanatory subgraphs from the input cell graphs.

%consider 1. diverse numbers of nuclei in different cell graphs and generate the explanatory subgraph at the same level of sparsity; 

%The separability scores are proposed to evaluate the inter-class separability of the nuclei attributes based on pathological concepts \cite{jaume2021quantifying}. The intuition is that the nuclei with different classes, which are more important for the predictions, are easier to separate. However, it fails to consider: 1. diverse numbers of nuclei in different cell graphs; 2. the separability of the explanatory subgraphs; 3. the separability of the complementary subgraphs. Therefore, we propose to compute the separability scores with the aggregated attributes. 

\textbf{Subgraph-level Attribute Aggregation:} 
Suppose we explain the prediction of a cell graph, where the nodes and edges denote nuclei and cellular interactions.
Given the explanatory subgraph $G_{e}$ and the complementary subgraph $G_{e\backslash}$, we aggregate their nuclei attributes $a\in \mathcal{A}$ of $G_{e}$ and $G_{e\backslash}$ via weighted summation:
\begin{equation}
\begin{aligned}
a_{G_{e}}^{k} = \sum_{i\in \mathcal{E}}M_{i}a_{i}/ \sum_{i\in \mathcal{E}} M_{i},
a_{G_{e\backslash}}^{k} = \sum_{i\in \mathcal{E\backslash}}M_{i}a_{i}/ \sum_{i\in \mathcal{E\backslash}} M_{i}
\end{aligned}
\label{attribute-agg}
\end{equation}
where $\mathcal{E}$ is the set of the top $k\%$ important nuclei in the input graph $G$ and $k$ is the sparsity of $G_{e}$. $M_{i}$ denotes the importance of the i-th nucleus in $\mathcal{E}$. $a_{G_{e}}^{k}$ and $a_{G_{e\backslash}}^{k}$ are the aggregated attributed for $G_{e}$ and $G_{e\backslash}$.
%Similarly, we replace $\mathcal{E}$ in Eq.~\ref{attribute-agg} with the complement set of nuclei and obtain the aggregated attributes $a_{G_{e\backslash}}^{k}$ of $G_{e\backslash}$. 

\textbf{Histogram Construction:}
For the cancer subtype $t\in \mathcal{T}$ and $k\in\mathcal{K}$, we construct the histograms of $a_{G_{e}}^{k}$ and $a_{G_{e\backslash}}^{k}$, denoted as $H_{t}^{k}(a_{G_{e}})$ and $H_{t}^{k}(a_{G_{e\backslash}})$. Then, we transform the histograms into the probability density functions.

\textbf{Separability Scores:}
The explanatory subgraphs encourage the inter-class separability of $H_{t}^{k}(a_{G_{e}})$ while reduce inter-class separability of $H_{t}^{k}(a_{G_{e\backslash}})$. Hence, we calculate the Wassertein distance between the probability density functions to evaluate the pair-wise class separability. 

We first compute the predictive separability scores $S$ of $G_{e}$. 
Given a class pair $(t_{x},t_{y})$, we compute the Wassertein distance between  $H_{t_{x}}^{k}(a_{G_{e}})$ and $H_{t_{y}}^{k}(a_{G_{e}})$. Then, we average the distances of attributes $a\in\mathcal{A}_{c}$ which belong to the same pathological concepts $c\in\mathcal{C}$, and obtain the concept distance $d_{c}^{k}(t_{x}, t_{y})$ under the sparsity $k$. Moreover, we compute the AUC of $d_{c}^{k}(t_{x}, t_{y})$ across different $k$, which is denoted as $D_{c}(t_{x}, t_{y})$. The maximal and average pair-wise metrics of $S$ are calculated as follows:

\begin{equation}
\begin{aligned}
&S_{max}(t_{x}, t_{y})=\max_{c\in \mathcal{C}}D_{c}(t_{x}, t_{y})\\
&S_{ave}(t_{x}, t_{y})=\frac{1}{|\mathcal{C}|}\sum_{c\in\mathcal{C}}D_{c}(t_{x}, t_{y})\\
\end{aligned}
\label{all-separability-score}
\end{equation}
 To evaluate the separability across all the class pairs, we sum up the pair-wise metrics for the aggregated metrics: $S_{max}$ and $S_{ave}$. We further consider the risk $R(t_{x}, t_{y})$ of misclassifying a sample of class $t_{x}$ into $t_{y}$ and vice versa, leading to the risk-aggregated metrics: $S_{max,R}$, $S_{ave,R}$. We empirically set the risks following \cite{jaume2021quantifying}. Thus, the predictive separability scores $S$ of $G_{e}$ consist of 1. the maximal and average pair-wise metrics: $S_{max}(t_{x}, t_{y})$ and $S_{ave}(t_{x}, t_{y})$; 2. the aggregated metrics: $S_{max}$ and $S_{ave}$; 3. the risk-aggregated metrics: $S_{max,R}$, $S_{ave,R}$.
 
Following the steps of calculating the predictive separability scores, we further compute the counterfactual separability scores $C$ of $G_{e\backslash}$.
Similarly, the counterfactual separability scores $C$ consist of 1. the maximal and average pair-wise metrics: $C_{max}(t_{x}, t_{y})$ and $C_{ave}(t_{x}, t_{y})$; 2. the aggregated metrics: $C_{max}$ and $C_{ave}$; 3. the risk-aggregated metrics: $C_{max,R}$, $C_{ave,R}$. Please refer to Supplementary Materials for more details.

\section{Experiments}

\subsection{Evaluation Metrics}
%Most evaluation metrics on explainers requires access to the output of the classifiers, which is model-dependent. Thus, we propose a novel set of metrics based on instance-level separability for a model-independent and pathologically aligned evaluation. 

\textbf{Fidelity:} The fidelity scores evaluate how the explanations are faithful to the GNN model \cite{yuan2021explainability}. We define the Fidelity+ score as the agreement between the original prediction and predictions of the explanatory subgraphs. The Fidelity- score is the agreement between the original predictions and the predictions after removing the explanatory subgraphs.
Let $y_{i}$ be the prediction of the i-th graph. $\hat{y}_{i}^{G_{e}}$ and $\hat{y}_{i}^{G_{e\backslash}}$ are the predictions of the explanatory and complementary subgraph of the i-th graph. The fidelity scores are computed as: $Fidelity+=\frac{1}{N} \sum_{i=1}^{N}\mathds{1}(y_{i}=\hat{y}_{i}^{G_{e}}),Fidelity-=\frac{1}{N} \sum_{i=1}^{N}\mathds{1}(y_{i}=\hat{y}_{i}^{G_{e\backslash}})$.
%\begin{equation}
%\begin{aligned}
%Fidelity^{+}&=\frac{1}{N} \sum_{i=1}^{N}\mathds{1}(y_{i}=\hat{y}_{i}^{G_{e}})\\
%Fidelity^{-}&=\frac{1}{N} \sum_{i=1}^{N}\mathds{1}(y_{i}=\hat{y}_{i}^{G_{e\backslash}})
%\end{aligned}
%\label{fid-positive}
%\end{equation}
Here $\mathds{1}(y_{i}=\hat{y}_{i}^{G_{e}})$ is the indicator function which outputs 1 if $y_{i}=\hat{y}_{i}^{G_{e}}$ and 0 otherwise.  Following \cite{jaume2021quantifying}, we filter out the misclassified samples.
For a fair comparison, the fidelity scores should be evaluated at the same level of sparsity, which is defined as the fraction of nodes in the explanatory subgraph: $k=|G_{e}|/|G|$. We set $k\in\{1\%,2\%,\cdots,10\%\}$ in the experiment.

\textbf{Separability:} 
 \begin{figure}[t]
\begin{center}
%\framebox[4.0in]{$\;$}
\centerline{\includegraphics[width=1.0\columnwidth]{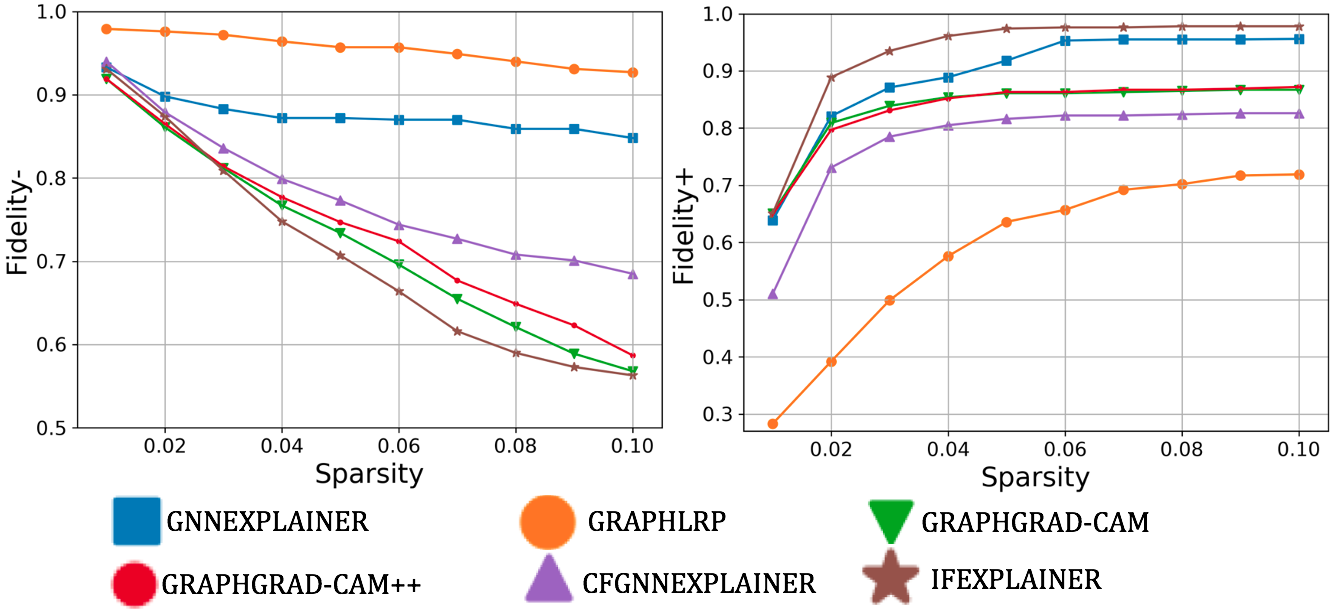}}
\vspace{-1.1cm}
\end{center}
\caption{A comparison on the fidelity scores of different explainers. IFEXPLAINER achieves best Fidelity-($\downarrow$) and Fidelity+($\uparrow$) on most sparsity scores.}
\label{fidelity}
\vspace{-0.6cm}
\end{figure}
We further evaluate the separability scores of the explanations introduced in Section~\ref{section-sep}. 
Specifically, the higher predictive separability scores $S$ indicate that the explanatory subgraphs $G_{e}$ are easier to classify. Hence, $G_{e}$ contains necessarily important information for the prediction. Meanwhile, the lower counterfactual separability scores $C$ show that the explanatory subgraphs $G_{e\backslash}$ are difficult to classify, leading to a different prediction.

\subsection{Dataset and Settings}
We employ the BReAst Cancer Subtyping (BRACS) dataset \cite{pati2020hact} for the experiments. It consists of 4931 RoIs at $40\times$ magnification sampled from 325 H\&E stained breast carcinoma whole-slides, and each RoI contains 1468 nuclei on average. 
We transform the RoIs into cell graphs following \cite{jaume2021histocartography}.
The labels of the cell graphs are Benign (B): normal, benign and usual ductal hyperplasia; Atypical (A):  flat epithelial atypia and atypical ductal hyperplasia; and Malignant (M): ductual carcinoma \textit{in situ} and invasive\cite{jaume2021quantifying}. 
The training, validation and test sets contain 3162, 602 and 626 cell graphs. 

We explain the pre-trained GNN which predicts the breast cancer subtypes of the cell graphs. It achieves 74.2\% weighted F1-score on the test set. We set $\beta=0.05$ and $\gamma=0.1$ for training IFEXPLAINER. Please refer to Supplementary Materials for more details on the experiments of hyperparameters.
%We transform the RoIs into cell graphs and employ a pretrained GNN classifier describe in Section~\ref{gnn-classifier} for classification. We use the same pretrained model in \cite{jaume2021quantifying}, which achieves 74.2\% weighted F1-score on the test set. 

\subsection{Baselines}
\textbf{GRAPHLRP \cite{schwarzenberg2019layerwise}:} 
GRAPHLRP outputs the importance score of each node in graph classification based on Layer-wise Relevance Propagation (LRP) \cite{bach2015pixel}. We employ GRAPHLRP to identify the importance scores of important nuclei in breast cancer subtyping. 

\textbf{GRAPHGRAD-CAM:} 
Based on GRAD-CAM \cite{selvaraju2017grad}, 
GRAPHGRAD-CAM computes the gradients w.r.t different GNN layers to output the node importance scores for the explanations of GNNs \cite{pope2019explainability}. 

\textbf{GRAPHGRAD-CAM++:} 
GRAPHGRAD-CAM++ considers the spatial contribution in the computation of the weitghted gradients for the improved node importance scores. We follow the implementation in \cite{jaume2021quantifying} to obtain the nuclei importance score.

\textbf{GNNEXPLAINER \cite{ying2019gnnexplainer,jaume2020towards}:} 
GNNEXPLAINER maximizes the mutual information between the distribution of the potential explanatory subgraphs and the original predictions for the explanations.

\textbf{CFGNNEXPLAINER \cite{lucic2021cf}:} 
CFGNNEXPLAINER recognizes a minimal subgraph of the input, if removed, would lead to a drastic change in the prediction.

\subsection{Performance}
We employ IFEXPLAINER and the baseline methods to interpret the predictions of the GNN for breast cancer subtyping. 
\begin{figure*}[t]
\begin{center}
%\framebox[4.0in]{$\;$}
\centerline{\includegraphics[width=2.0\columnwidth]{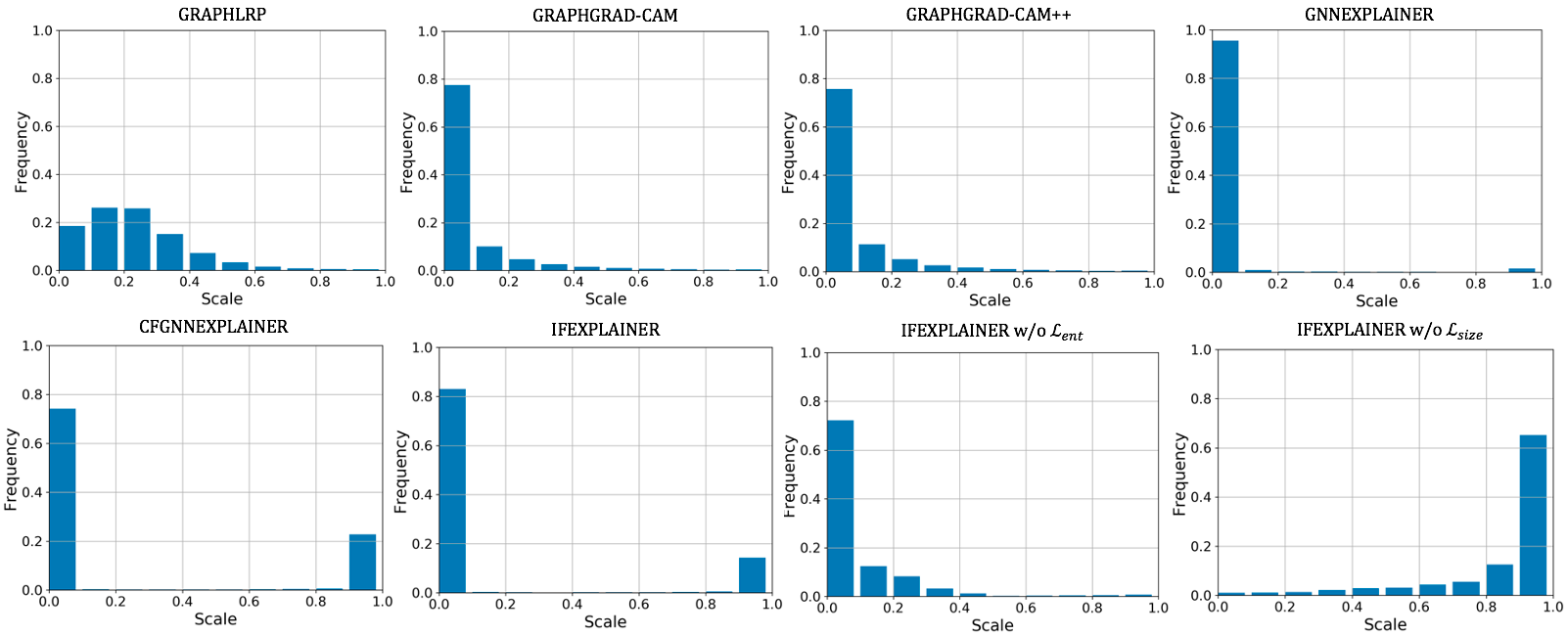}}
\vspace{-1.1cm}
\end{center}
\caption{The distributions of nuclei importance scores generated by different explainers. } 
\label{scale}
\vspace{-0.7cm}
\end{figure*}
As shown in Table~\ref{separability}, IFEXPLAINER achieves higher predictive separability scores $S$ over the baselines on most pair-wise metrics and all the aggregated metrics.
This shows that IFEXPLAINER generates explanatory subgraphs with higher inter-class separability, and thus are most important to the predictions.
Moreover, IFEXPLAINER outperforms the baselines with lower counterfactual separability scores $C$ on most pair-wise and aggregated metrics.
Hence, the complementary subgraphs obtained by removing the explanatory subgraphs are least separable and most likely cause the prediction changes.
Noticeably, the predictive separability scores are consistently higher than the corresponding counterfactual separability scores for IFEXPLAINER.
This shows that IFEXPLAINER uncovers the most important nuclei for the cancer subtype.
However, most baseline methods have a higher $C_{ave}$ than $S_{ave}$ for the class pair (B v.s. A).
This shows that the explanatory subgraphs produced by IFEXPLAINER identify most necessarily and sufficiently important nuclei for the breast cancer subtyping.

We further compare the fidelity scores of different methods. 
As the sparsity increases, the Fidelity- score of IFEXPLAINER declines drastically in Figure~\ref{fidelity}. 
This shows that the removal of the explanation is most likely to cause the prediction change. 
IFEXPLAINER also outperforms the baselines on the Fidelity+ scores, which indicates the produced explanation is most important to the prediction.
Noticeably, the Fidelity- score of GNNEXPLAINER declines slowly as the sparsity increases. Thus, GNNEXPLAINER ignores some predictive nuclei for the prediction.

\subsection{Visualization}
We visualize the nuclei importance scores generated by different explanation methods. As shown in Figure~\ref{show-cells}, IFEXPLAINER generates almost binarized importance scores and emphasizes the nuclei leading to the predictions. Similarly, GNNEXPLAINER and CFGNNEXPLAINER also generate binarized node importance scores. However, the results of GNNEXPLAINER are less interpretable since almost all nuclei are categorized as unimportant to the breast cancer subtypes. 
And CFGNNEXPLAINER only attaches importance to a small portion of nuclei that are relevant to the predictions.
Noticeably, GRAPHLRP and gradient-based methods such as GRAPHGRAD-CAM and GRAPHGRAD-CAM++ consistently highlight similar regions. 
Unlike IFEXPLAINER, their importance maps are less pathologist-intelligible since most nuclei are attached to low importance scores. 

We further analyze the distributions of nuclei importance scores produced by different methods. Ideally, the importance scores are binary values, indicating whether the nucleus is informative for the prediction. In practice, the explanation models either evaluate the nuclei importance heuristically, or optimize a differentiable node mask. Hence, the output importance scores are continuous values between [0,1]. However, the new semantic meaning or noises introduced by the continuous importance scores prevent the pathologists from reasoning the explainability of GNNs. Thus, it is important to generate binarized importance scores.
As shown in Figure~\ref{scale}, GRAPHLRP and the gradient-based methods generate continuous importance scores and assign low importance to most nuclei. GNNEXPLAINER attaches low importance to most nuclei, which is less interpretable. 
Differently, both CFGNNEXPLAINER and IFEXPLAINER generate almost binarized importance scores. 
However, IFEXPLAINER outperforms CFGNNEXPLAINER in qualitative and quantitative studies since the explanation produced by CFGNNEXPLAINER is not necessarily important to the prediction.

%generates pathologist-interpretable explanations 

\subsection{Ablation Study}
To evaluate the influences of $\mathcal{L}_{size}$ and $\mathcal{L}_{ent}$ on the produced explanations, we individually remove these terms in the objective, which leads to two variant models, namely IFEXPLAINER w/o $\mathcal{L}_{size}$ and IFEXPLAINER w/o $\mathcal{L}_{ent}$. We first evaluate the generated explanations of these methods in terms of separability scores. As shown in Table~\ref{separability}, IFEXPLAINER outperforms the variants on most pair-wise and aggregated separability scores, which shows the efficacy of $\mathcal{L}_{size}$ and $\mathcal{L}_{ent}$. Moreover, we study the distributions of nuclei importance scores of these variants. As shown in Figure~\ref{fidelity}, IFEXPLAINER w/o $\mathcal{L}_{size}$ attaches high importance to most nuclei as there is no constraint on the sizes of the explanatory subgraphs. And IFEXPLAINER w/o $\mathcal{L}_{ent}$ can not generate binary importance scores without the entropy constraint.

\section{Discussions}
\textbf{Potential Negative Impacts:} Our explainer is designed for improved transparency in GNN-based clinical decisions. The concern is that this technique, if not adequately used under administration, may lead to the privacy leakage. 

\textbf{Limitations:} IFEXPLAINER generates the customized explanation to the prediction of each biological instance without access to the node embeddings inside the black-box GNNs. Thus, it can be easily adapted to interpret various GNN models in digital pathology tasks.
However, this also limits the usage of our model in inductive settings. We leave it in our future work.

\section{Conclusion}
In this work, we emphasize the necessary and sufficient nature of the explanations and propose a novel explainer of GNN in digital pathology, namely IFEXPLAINER. IFEXPLAINER exploits the information flows from different input substructures to the prediction with the notion of $f$-Information, and generates faithful explanation to the GNN's prediction. 
We also propose a set of evaluation metrics based on subgraph-level separability for the comprehensive assessments to the explanation models.
Experimental results on the BRACS dataset show the superior performance of the proposed method.
%-------------------------------------------------------------------------

%%%%%%%%% REFERENCES
{\small
\bibliographystyle{plain}
\bibliography{egbib}
}

\end{document}